\documentclass{article}
\usepackage{ijcai18}
\usepackage{times}
\usepackage{setspace}
\usepackage{soul}
\usepackage{url}
\usepackage[utf8]{inputenc}

\usepackage{color}
\usepackage{comment}

\usepackage{booktabs}
\usepackage{amsfonts}
\usepackage{amsmath}
\usepackage{amssymb}
\usepackage{makecell}
\usepackage{subcaption}
\usepackage{multirow}
\usepackage{pbox}

\usepackage[flushleft]{threeparttable}

\usepackage{graphicx,xspace}
\usepackage{epstopdf}
\usepackage{stfloats}
\usepackage[font=small]{caption}
\usepackage[ruled, vlined, linesnumbered, resetcount, noend]{algorithm2e}
\usepackage{array}
\newcommand{\modelnamens}{\texttt{KDCoE}}
\newcommand{\modelname}{{\modelnamens}\ }
\newcolumntype{?}{!{\vrule width 1pt}}
\def\st#1{~}
\def\inv{\vspace{-0.0cm}}

\begin{document}
%
\title{Co-training Embeddings of Knowledge Graphs and Entity Descriptions\\ for Cross-lingual Entity Alignment}
\author{Muhao Chen$^1$, Yingtao Tian$^2$, Kai-Wei Chang$^1$, Steven Skiena$^2$ {\normalfont and} Carlo Zaniolo$^1$\\
$^1$Department of Computer Science, University of California, Los Angeles\\
$^2$Department of Computer Science, Stony Brook University\\
\{muhaochen, kw2c, zaniolo\}@cs.ucla.edu; \{yittian, skiena\}@cs.stonybrook.edu\\
}
\maketitle
\begin{abstract}
Multilingual knowledge graph (KG) embeddings
provide latent semantic
representations of entities and
structured knowledge with cross-lingual inferences,
which benefit various knowledge-driven cross-lingual NLP tasks.
However, precisely learning such cross-lingual inferences is usually hindered by the low coverage of entity alignment in many KGs.
Since many multilingual KGs also provide literal descriptions of entities,
in this paper, we introduce an embedding-based approach which leverages a weakly aligned multilingual KG for semi-supervised cross-lingual learning using entity descriptions.
Our approach performs co-training of two embedding models, i.e. a multilingual KG embedding model and a multilingual literal description embedding model.
The models are trained on a large Wikipedia-based trilingual dataset where most entity alignment is unknown to training.
Experimental results show that the performance of the proposed approach on the entity alignment task improves at each iteration of co-training, and eventually reaches a stage at which it significantly surpasses previous approaches.
We also show that our approach has promising abilities for zero-shot entity alignment, and cross-lingual KG completion.
\end{abstract}

\section{Introduction}
\begin{figure}[th!]
  \centering
  \includegraphics[width=0.96\columnwidth]{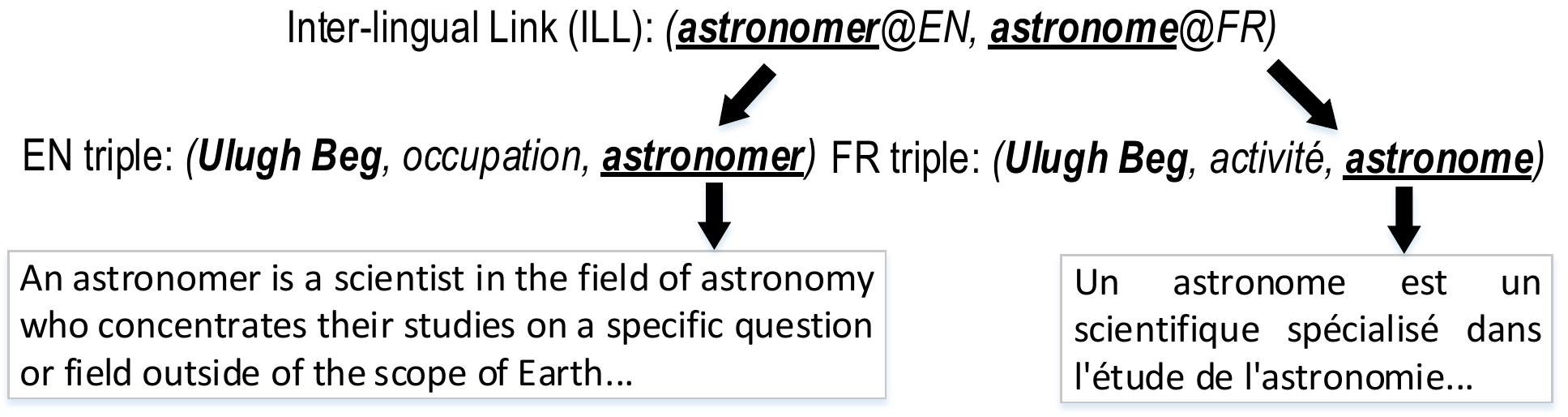}\\
  \caption{A simple example which shows triples, an ILL, and entity descriptions in a multilingual KG (DBpedia). The French description for \emph{astronome} means \emph{an astronomer is a scientist specialized in the study of astronomy}, which contains much fewer content details than the English description for \emph{astronomer}.}~\label{fig:desc}
\end{figure}

Multilingual knowledge bases (KBs) such as \mbox{DBpedia} \cite{lehmann2015dbpedia}, ConceptNet~\cite{speer2017conceptnet},
and Yago~\cite{mahdisoltani2014yago3} constitute crucial sources of knowledge for AI-related applications.
These KBs
store knowledge graphs (KGs) that represent two aspects of structured knowledge:
(1) the {\em monolingual knowledge} that models relational facts of entities as triples,
(2) and the {\em cross-lingual knowledge} that synchronizes monolingual knowledge among multiple human languages
(see Fig.~\ref{fig:desc}).
In addition to those, many KGs also store 
literal descriptions of entities in different languages~\cite{xie2016desc,lehmann2015dbpedia}.

Embedding models for KGs have been extensively studied in the past few years.
These models aim at characterizing entities in low-dimensional embedding spaces,
and supporting relational inferences for entity embeddings via simple vector algebra.
Hence, they provide efficient and versatile methods to incorporate the symbolic knowledge of KGs into machine learning.
Models of this kind have been widely applied to NLP-related tasks, such as 
relation extraction \cite{wang2014knowledge}, ontology population \cite{chen2018onto}, 
question answering (QA)~\cite{bordes2014open}, dialogue agents \cite{he2017learning}, and visual semantic labeling \cite{fangobject2017}.\par


Recently, embedding models are leveraged to \mbox{connect} KG structures of multiple languages~\cite{chen2017multigraph,chen2017affine,sun2017cross,zhu2017iterative}.
Emerging of such approaches is significant, inasmuch as they extend the inferences of KG embeddings to a multilingual scenario, and seek to benefit cross-lingual NLP tasks such as knowledge alignment, cross-lingual QA and machine translation. 
While such embeddings are generic and beneficial,
it remains very challenging for corresponding approaches to precisely capture the cross-lingual inferences.
The challenge is that the cross-lingual knowledge, which is typically formed as \emph{inter-lingual links} (ILLs) that match cross-lingual counterparts of entities, is usually far from complete.
In fact, ILLs cover less than 20\% of the entities even in the most successful Wikipedia-based KBs. 
Hence, the lack of supervision by cross-lingual knowledge easily hinders the quality of cross-lingual inferences, 
which affects even more significantly when each language version of KG scales up and becomes inconsistent in contents and density.\par

While existing embedding models solely rely on the structured knowledge for cross-lingual learning, it would be promising to enhance the corresponding learning process with the literal descriptions of entities that are stored in many KGs~\cite{xie2016desc,lehmann2015dbpedia,mahdisoltani2014yago3}.
These descriptions comprise an alternative view of entities that potentially bridges two languages, since the descriptions of an entity 
in different languages often share a lot of semantic information.
However, it is non-trivial to characterize and utilize such information for cross-lingual learning,
as this requires the model to learn to match descriptions across different languages with inadequate labels, while conquering the inconsistency of literals in content details, grammars, and word orders (as shown in Fig.~\ref{fig:desc}).
Moreover, 
aggregating semantic relatedness of descriptions from words of different languages is another challenge.\par

To address 
these issues, we propose a novel co-training-based approach \modelname to enhance the semi-supervised learning of multilingual KG embeddings.
\modelname iteratively trains two component embedding models on multilingual KG structures and entity descriptions respectively.
A KG embedding model jointly trains a translational knowledge model with a linear-transformation-based alignment model to encode the KG structure.
A description embedding model employs an attentive gated recurrent unit encoder (AGRU) and multilingual word embeddings to characterize multilingual entity descriptions, and is trained to collocate the embeddings of cross-lingual counterparts. 
The co-training is processed on a large Wikipedia-based trilingual KG, for which a very \mbox{small} portion of ILLs is used for training.
During each iteration of co-training, both models alternately propose a set of most confident new ILLs to strengthen the supervision of cross-lingual learning, which leads to gradually improved accuracy on cross-lingual inferences.
Experimental results on entity alignment confirms the effectiveness of \modelname that significantly outperforms previous models, while those results on zero-shot alignment and cross-lingual KG completion also show wider usability of our approach.\par

The rest of the paper is organized as follows. We first discuss the related work, and then introduce our approach in the section that follows. After that we present the experimental results, and conclude the paper in the last section.

\newcommand{\stitle}[1]{\vspace{0.3ex}\noindent{\bf #1}}
\inv
\section{Related Work} \label{sec:related}
We discuss three lines of works that are relevant to this paper.

\stitle{Monolingual KG Embeddings.}
KG embeddings are first explored in the monolingual scenario.
The past half decade has seen much popularity on translational models,
which mostly follow the forerunner TransE~\cite{bordes2013translating} to capture a triple $(h, r, t)$ as a translation $\mathbf{r}$ between two entity embeddings $\mathbf{h}$, $\mathbf{t}$.
Later works such as TransH~\cite{wang2014knowledge}, TransR~\cite{lin2015learning}, TransD~\cite{ji2015knowledge}, and TransA~\cite{jia2016locally} differentiate such translations in separated spaces using different forms of relation-specific projections.
Models of this family preserve well the KG structures in the embedding spaces regardless of their simplicity, and offer promising performance on KG completion and relation extraction tasks.
In \cite{xie2016desc} TransE is trained jointly with a convolutional neural network (CNN) to predict entities based on their descriptions.
In addition to them, recent works also introduce 
successful non-translational models, such as DistMult \cite{yang2015embedding} and HolE \cite{nickel2016holographic} that adopt dot product and circular correlation respectively, and neural models such as ConvE \cite{dettmers2018convolutional}.
These models perform comparably to or even better than translational models at the cost of model complexity.\par
\stitle{Multilingual KG Embeddings.} More recent work extends embedding models to multilingual learning on KGs.
One representative work is MTransE \cite{chen2017multigraph}.
\mbox{MTransE} connects monolingual models with a jointly trained alignment model, for which three aligment techniques are employed, i.e., axis calibration that adjusts embedding \mbox{spaces} to collocate cross-lingual counterparts (MTransE-AC), vector translation (MTransE-TV), and linear transformations across embedding spaces (MTransE-LT) for \mbox{different} languages.
MTransE-LT thereof achieves the best performance on knowledge alignment tasks.
JAPE is introduced in \cite{sun2017cross}
to strengthen the cross-lingual learning of MTransE-AC based on the similarity of entity attributes.
This model performs well on KBs that provide numerical entity attributes, though such attributes are not generally available in many KBs.
Another relevant model ITransE \cite{zhu2017iterative} incorporates self-training into a hard-alignment version of MTransE-AC.
ITransE is used to align entities across monolingual KGs with coherent vocabularies and triples, but we find it does not \mbox{adapt} well to the inconsistent multilingual scenario.
Note that off-line multilingual word embedding models, including LM \cite{mikolov2013exploiting}, CCA \cite{faruqui2014improving}, and orthogonal-transformation-based OT \cite{xing2015normalized} can also be extended to KGs, but are outperformed by MTransE on cross-lingual tasks.\par
\stitle{Co-training.} 
Co-training combines multiple 
models to learn on different views of the data in the training process, in which all participating models 
take turn in suggesting more labels on unlabeled data to enhance the supervision.
This technique is widely used in semi-supervised learning tasks, such as sentiment classification on bilingual corpora with incomplete labels~\cite{wan2009co}, collaborative filtering in recommender systems with multiple user views~\cite{zhang2014addressing}, and semantic role labeling based on the semantic and syntactic views of documents~\cite{do2016facing}.
Our work conducts co-training on two views of the multilingual KG, i.e. structures and literal descriptions,
which to the best of our knowledge, is the first work that incorporates co-training into embedding learning, as well as knowledge alignment tasks.

\def\kb{\mathit{KB}}
\def\lang{\mathcal{L}}
\def\bhline{\specialrule{.2em}{0em}{0em}}
\newcommand{\bigO}[1]{{\rm O} (#1)\xspace}

\section{Modeling}
We first provide the definition of multilingual KGs.
In a \mbox{KB}, $\lang$ denotes the set of languages, and $\lang^2$ 
unordered language pairs.
$G_{L}$ is the language-specific KG of each language $L \in \lang$.
$E_{L}$ and $R_{L}$ respectively denote the corresponding vocabularies of entities and relations.
$T = (h,r,t)$ denotes a triple in $G_{L}$ such that $h, t \in E_{L}$ and $r \in R_{L}$.
Boldfaced $\mathbf{h}$, $\mathbf{r}$, $\mathbf{t}$
represent the embedding vectors of head $h$, relation $r$, and tail $t$ respectively. For a language pair
$(L_1, L_2) \in \lang^2$,
$I(L_1, L_2)$ denotes a set of ILLs that align entities
between $L_1$ and $L_2$, such that $e_1\in E_{L_1}$ and $e_2\in E_{L_2}$ for each \mbox{ILL} $(e_1, e_2)\in I(L_1, L_2)$.
We assume the entity pairs have a 1-to-1 mapping and it is specified in $I(L_1, L_2)$.
This assumption is congruent to the design of mainstream KGs~\cite{lehmann2015dbpedia}. 
Besides the above structured knowledge, we use $D_L$ to denote the literal descriptions of entities in language $L$.
A description $d_e\in D_L$ describes an entity $e\in E_L$ with a sequence of words from the word vocabulary $W_L$, i.e. $d_e=\{w_1, w_2, ..., w_l\}$.
\par

\modelname conducts iterative co-training of two components, 
i.e. the multilingual KG embedding model (KGEM) and the multilingual description embedding model (DEM),
which capture embeddings with cross-lingual inferences for structured knowledge and entity descriptions respectively.
During co-training, both components are trained in turns to propose new ILLs with high confidence,
which 
populate the training set and become visible to future turns of training.
We define the model on a pair of languages from $\lang^2$ for which the ILLs are provided.
For a KB with more than two languages, multiple models that bridge different languages compose the solution \emph{w.l.o.g.}
In the following subsections, we use a language pair $(L_i, L_j)\in \lang^2$ to describe the definition of the model components and the entire learning process.

\subsection{Multilingual KG Embeddings}
The KGEM consists of two components that learn on the two facets of structured knowledge.\par
A knowledge model is learnt to preserve entities and relations of each language in a separated embedding space.
Specifically, for each participating language $L$, a dedicated $k_1$-dimensional embedding space $\mathbb{R}_{L}^{k_1}$ is assigned for vectors of $ E_{L}$ and $ R_{L}$. 
Like previous work~\cite{chen2017multigraph,sun2017cross,zhu2017iterative}, we adopt the basic translational
method of TransE for each involved language, which benefits the cross-lingual tasks with uniform representations of entities in different contexts of relations.
The corresponding objective function is given as the following hinge loss,
\begin{equation*}
S_K = \sum_{L \in \{L_i, L_j\}} \sum_{(h,r,t) \in G_L\wedge(\hat{h},r,\hat{t})\notin G_L} [f_r(h,t)-f_r(\hat{h},\hat{t})+\gamma]_+
\end{equation*}
for which $f_r(h,t)=\left \| \mathbf{h}+\mathbf{r}-\mathbf{t} \right \|_2$ is the dissimilarity measure of a triple $(h,r,t)$, $\gamma$ is a positive margin, $[x]_+$ denotes the positive part of $x$ (i.e. $\max(x, 0)$), and
$(\hat{h},r,\hat{t})$ is a Bernoulli negative-sampled triple~\cite{wang2014knowledge} by substituting either $h$ or $t$ in $(h,r,t)$.\par
On top of that, an alignment model jointly captures cross-lingual inferences across language-specific embedding \mbox{spaces}.
As previously mentioned, various alignment techniques have been adopted by previous models, among which we choose the linear-transformation-based technique, as we find it offers the best performance in modeling cross-lingual inferences.
Hence, the objective function is given as below.
\begin{equation*}
S_{A} = \sum_{(e, e') \in I(L_{i}, L_{j})} \left \| \mathbf{M}_{ij}\mathbf{e}-\mathbf{e}' \right \|_2
\end{equation*}
$\mathbf{M}_{ij}$ thereof is a $k_1 \times k_1$ matrix that serves as a linear transformation on entity vectors from $L_i$ to $L_j$.\par

The objective of the KGEM component is to minimize $S_{KG}=S_K+\alpha S_A$, for which $\alpha$ is a positive hyperparameter.
Conceptually, KGEM is equivalent to a modification of MTransE-LT, where the alignment model is refined from the triple level to the entity level.
Meanwhile, the linear transformation $\mathbf{M}_{ij}$ also applies to relation embeddings.
Consider two cross-lingual counterparts of triples $(h,r,t)$, $(h',r',t')$, since $S_K$ requires $\mathbf{h}+\mathbf{r}\approx \mathbf{t}$ and $\mathbf{h}'+\mathbf{r}'\approx \mathbf{t}'$, it is easy to get $\mathbf{M}_{ij}\mathbf{r} \approx \mathbf{r'}$ from $\mathbf{M}_{ij}(\mathbf{t}-\mathbf{h}) \approx \mathbf{t}'-\mathbf{h}'$.
It is noteworthy that, other techniques such as DistMult and HolE may be employed for the knowledge model as well, but we leave them to future work for two reasons: (1) to facilitate the direct comparison with previous works; (2) these techniques disable the cross-lingual inferences on relations.\par

Like many KG embedding models~\cite{bordes2013translating,bordes2014open,yang2015embedding,chen2017multigraph}, we constrain the $l_2$-norms of entity embeddings to be 1 during the learning process.
We notice that some other works~\cite{wang2014knowledge,lin2015learning} constrain such norms to be less than 1 instead, which we find however, do not prevent the optimization from a trivial solution where all vectors collapse towards zero, hence largely impairing the quality of embeddings.

\subsection{Multilingual Entity Description Embeddings}~\label{sec:KGEM}
The DEM learns in two stages.
An attentive gated recurrent unit encoder (AGRU) is used to encode the multilingual entity descriptions.
On top of that, DEM is trained to collocate the description embeddings of cross-lingual counterparts.\par
\stitle{Gated Recurrent Unit}. The gated recurrent unit (GRU) has been popular in sentence (sequence) encoders in recent works \cite{jozefowicz2015empirical},
which consists of two types of gates to track the state of sequences without using separated memory cells, i.e. the reset gate $\mathbf{g}_t$ and the update gate $\mathbf{z}_t$.
Given the vector representation $\mathbf{x}_t$ of an incoming item $x_t$ from the sequence, GRU updates the current state $\mathbf{s}_t$ as a linear interpolation between the previous state $\mathbf{s}_{t-1}$ and the candidate state $\tilde{\mathbf{s}_t}$ of the new item $x_t$, which is calculated as below.
\begin{equation*}
\mathbf{s}_t=\mathbf{z}_t\odot \tilde{\mathbf{s}_t}+(1-\mathbf{z}_t)\odot \mathbf{s}_{t-1}
\end{equation*}
The update gate $\mathbf{z}_t$ that balances between the information of the previous sequence and the new item is updated as below,
\begin{equation*}
\mathbf{z}_t=\sigma\left (\mathbf{M}_z\mathbf{x}_t+\mathbf{N}_z \mathbf{s}_{t-1} + \mathbf{b}_z\right )
\end{equation*}
where $\mathbf{M}_z$ and $\mathbf{N}_z$ are two weight matrices, $\mathbf{b}_z$ is a bias vector, and $\sigma$ is the sigmoid function.
The candidate state $\tilde{\mathbf{s}_t}$ is calculated similarly to those in a traditional recurrent unit as below, where $\mathbf{M}_s$ and $\mathbf{N}_s$ are two weight matrices, and $\mathbf{b}_s$ is a bias vector,
\begin{equation*}
\tilde{\mathbf{s}_t}=\mathrm{tanh}\left (\mathbf{M}_s\mathbf{x}_t+\mathbf{g}_t\odot(\mathbf{N}_s \mathbf{s}_{t-1}) + \mathbf{b}_s\right )
\end{equation*}
The reset gate $\mathbf{g}_t$ thereof, controls how much the information from the past sequence contribute to the candidate state, and is updated as below,
\begin{equation*}
\mathbf{g}_t=\sigma\left (\mathbf{M}_g\mathbf{x}_t+\mathbf{N}_g \mathbf{s}_{t-1} + \mathbf{b}_g\right )
\end{equation*}
The above defines a GRU layer
which outputs a sequence of hidden state vectors given the input sequence $X$.
\par
\stitle{Self-attention.}
The motivation of importing self-attention is to extract the words that contain shared information across the multilingual descriptions of the same entity, where content details can be inconsistent.
Consider the two descriptions of Fig.~\ref{fig:desc}. We expect the encoder to highlight the sentence parts with the important shared information such as \emph{scientist} and \emph{scientifique sp\'ecialis\'e} as well as \emph{the field of astronomy} and \emph{l'astronomie}, but rather than inconsistent details such as \emph{specific question or field outside of the scope of Earth} in English.
We hence incorporate the self-attention defined in~\cite{kim2017structured} to the above GRU layer as below.
\begin{align*}
\begin{split}~\label{eq:attention}
&\mathbf{u}_t=\mathrm{tanh}\left (\mathbf{M}_a\mathbf{s}_t+\mathbf{b}_a\right )\\
&a_t=\frac{\mathrm{exp}\left (\mathbf{u}^\top_t\mathbf{x}_t\right )}{\sum_{x_i\in X}\mathrm{exp}\left (\mathbf{u}^\top_i\mathbf{x}_i\right )}\\
&\mathbf{v}_t=|X|a_t\mathbf{u}_t
\end{split}
\end{align*}
$\mathbf{u}_t$ thereof is a hidden representation of $\mathbf{s}_t$ from the GRU layer.
A normalized attention weight $a_t$ is calculated through a softmax function, which measures the importance of item $x_t$ in the GRU encoding of sequence $X$, and is applied to $\mathbf{u}_t$ to obtain the self-attention output $\mathbf{v}_t$.
Note that a coefficient $|X|$ (the length of the input sequence) is applied so as to keep $\mathbf{v}_t$ from losing the original scale.

\stitle{Multilingual Word Embeddings.}
To better reflect the semantic information of multilingual entity descriptions from the word level,
we use multilingual word embeddings that are capable of collocating similar words in different languages. 
%
In detail, we pre-train the cross-lingual Bilbowa \cite{gouws2015bilbowa} word embeddings on 
the cross-lingual parallel corpora Europarl v7 \cite{koehn2005europarl} 
and monolingual corpora of Wikipedia dump. 
After the pre-training, we fix the word embeddings to convert each entity description $d_e$ to a sequence of vectors to be fed into the description encoder.

\par
\stitle{Learning Objective.}
We utilize an encoder of two stacked attentive GRU layers to model the descriptions of both languages, which takes the description sequence $d_e$
and produces the embedding from the second-layer outputs.
In detail, we apply an affine layer to map the averaged second-layer outputs to a common embedding space for \mbox{descriptions}:
$\mathbf{d}_e=\mathrm{tanh}\left(\mathbf{M}_d \left( \frac{1}{|d_e|}\sum_{i=1}^{|d_e|}\mathbf{v}_i^{(2)} \right) +\mathbf{b}_d \right)$.
We use the same dimensionality (denoted as $k_2$) for the output vectors of the second GRU layer $\mathbf{v}_i^{(2)}$ and the description embeddings $\mathbf{d}_e$.
Like KG embeddings, we regularize each $\mathbf{d}_e$ as $\left \| \mathbf{d}_e \right \|_2=1$.\par
The learning objective of DEM is to maximize the log likelihood of each entity given its cross-lingual counterpart in terms of their description embeddings, which is realized by minimizing the following objective function,
\begin{align*}
\begin{split}
S_{D}&=\sum_{(e, e')\in I(L_i,L_j)}-LL_{1}-LL_{2}\\
&=\sum_{(e, e')\in I(L_i,L_j)}-\mathrm{log}\left (P(e|e')\right )-\mathrm{log}\left (P(e'|e)\right )
\end{split}
\end{align*}
Similar to \cite{mikolov2013word2vec}, we adopt negative sampling to obtain the following computationally efficient terms of approximation for $LL_{1}$ and $LL_{2}$, where $|B_d|$ is the batched sampling size, and $\mathrm{U}$ is the distribution of entities.
\small
\begin{align*}
\begin{split}
&LL_{1}=\mathrm{log}\sigma\left ( \mathbf{d}_e^\top \mathbf{d}_{e'} \right )+\sum_{k=1}^{|B_d|}\mathbb{E}_{e_k\sim \mathrm{U}\left ( e_k\in E_{L_i} \right )}\left [ \mathrm{log}\sigma\left (- \mathbf{d}_{e_k}^\top \mathbf{d}_{e'} \right ) \right ]\\
&LL_{2}=\mathrm{log}\sigma\left ( \mathbf{d}_e^\top \mathbf{d}_{e'} \right )+\sum_{k=1}^{|B_d|}\mathbb{E}_{e_k\sim \mathrm{U}\left ( e_k\in E_{L_j} \right )}\left [ \mathrm{log}\sigma\left (- \mathbf{d}_e^\top \mathbf{d}_{e_{k}} \right ) \right ]
\end{split}
\end{align*}
\normalsize
Through optimization of $S_{D}$, the encoder is trained towards the goal of maximizing the dot product of each description embedding $\mathbf{d}_e$ and that of its cross-lingual counterpart $\mathbf{d}_{e'}$, and decreasing the dot product of unrelated description embeddings.
Since description embeddings are regularized to unit vectors, this process is equivalent to minimizing the $l_2$-distance between each pair of cross-lingual counterparts (i.e. collocating).
To facilitate the sampling-based approximation, we use the stratified negative sharing technique~\cite{chen2017sampling}.
That is to say, we sample batches of ILLs into $B_d$.
Then based on the 1-to-1 mapping of ILLs, we select negative samples for each $e$ as all entities $e_k$ in the other language from $B_d$, except for the one that forms the ILL with $e$.\par

Note that we have also explored with other forms of description encoders.
Single-layer (an affine layer applied to averaged word embeddings) and CNN 
used in \cite{xie2016desc} to represent monolingual entity descriptions fail to accurately match cross-lingual counterparts by losing the sequential and attentive information.
Attentive LSTM encoders perform comparably to AGRU, but are more complex and require more computational resources for training.
Adopting bidirectional encoders hinders the performance of our tasks.\par
\subsection{Iterative Co-training}

{\LinesNotNumbered
\begin{algorithm}[t]
  \caption{Iterative co-training of \modelnamens.}~\label{alg:training}
  \scriptsize
  \KwIn{Graphs $G_{L_i}$, $G_{L_j}$, descriptions $D_{L_i}$, $D_{L_j}$, ILL training set $I_{tr}$, ILL validation set $I_{val}$, candidate entities without ILLs $\tilde{E_{L_i}}\in E_{L_i}$, $\tilde{E_{L_j}}\in E_{L_j}$, precision threshold $\tau$ on $I_{val}$ for selecting proposed ILLs.}
  \LinesNumbered
  \KwOut{parameters $\theta$ for KGEM and DEM}
   \While{Either \textup{KGEM} or \textup{DEM} does not propose more ILLs}{
      Reinitialize KGEM and DEM\;
      Train KGEM on $I_{tr}, G_{L_i}$, $G_{L_j}$ until $S_{KG}$ no longer improves on graphs and $I_{val}$\;
      Select max $l_2$ threshold $\delta_1$, for which the precision of the predictions $(e, \hat{e'})$ by KGEM on $I_{val}$ s.t. $\left \| \mathbf{M}_{ij}\mathbf{e} - \hat{\mathbf{e}'} \right \|_2 <\delta_1$ is higher than $\tau$\;
      \For{$e\in \tilde{E_{L_i}}$}{
        $\hat{\mathbf{e}'} \leftarrow \mathrm{NearestNeighbor}(\mathbf{M}_{ij}\mathbf{e}, L_j)$\tcc*[r]{NN in $L_j$.}
        \If {$\left \| \mathbf{M}_{ij}\mathbf{e} - \hat{\mathbf{e}'} \right \|_2 <\delta_1$} {
          $I_{tr} \leftarrow I_{tr} \cup \{(e, \hat{e'})\}$\tcc*[r]{Propose an ILL.}
          $\tilde{E_{L_i}} \leftarrow \tilde{E_{L_i}}-\{e\}$; $\tilde{E_{L_j}} \leftarrow \tilde{E_{L_j}}-\{\hat{e'}\}$\;
        }
      }
      Train DEM on $I_{tr}, D_{L_i}, D_{L_j}$ until $S_{D}$ no longer improves on $I_{val}$\;
      Select max $l_2$ threshold $\delta_2$, for which the precision of the predictions $(e, \hat{e'})$ by DEM on $I_{val}$ s.t. $\left \| \mathbf{d}_e - \mathbf{d}_{\hat{e'}} \right \|_2 <\delta_2$ is higher than $\tau$\;
      \For{$e\in \tilde{E_{L_i}}$}{
        $\mathbf{d}_{\hat{e'}} \leftarrow \mathrm{NearestNeighbor}(\mathbf{d}_e, L_j)$\tcc*[r]{NN in $L_j$.}
        \If {$\left \| \mathbf{d}_e - \mathbf{d}_{\hat{e'}} \right \|_2 <\delta_2$} {
          $I_{tr} \leftarrow I_{tr} \cup \{(e, \hat{e'})\}$\tcc*[r]{Propose an ILL.}
          $\tilde{E_{L_i}} \leftarrow \tilde{E_{L_i}}-\{e\}$; $\tilde{E_{L_j}} \leftarrow \tilde{E_{L_j}}-\{\hat{e'}\}$\;
        }
      }
   }
\end{algorithm}
}

The co-training of the two model components is conducted iteratively on the KG, where a small amount of ILLs is provided for training.
At each iteration, 
the component \mbox{models} alternately take turns of the train-and-propose process.
In each turn, the model is first initialized using orthogonal initialization, 
and optimized using SGD with early-stopping based on a small validation set of ILLs.
After training, that model predicts new ILLs for candidate entities that are not involved in any previous \mbox{ILL}.
Such a prediction is based on a distance-based strategy, where a new ILL sourced from $L_i$ is suggested by searching the \mbox{nearest} neighbor (NN) within the candidate space of $L_j$ from the transformed entity vector, or from the original description vector.
As lower $l_2$-distances imply more precise inferences of embeddings \cite{chen2017multigraph,zhu2017iterative,mikolov2013word2vec},
only the most confident predictions, for which the $l_2$-distance between the source and the NN falls within a certain threshold, are populated into the training set.
The $l_2$-distance threshold is selected to ensure the prediction precision on the validation set to be above $\tau$, so as to ensure a high estimated precision of proposed new ILLs.
Both components repeatedly conduct the above train-and-propose processes, therefore gradually enhance the supervision of cross-lingual learning for each other, until either of the two model components no longer proposes new ILLs.
The detailed co-training procedure of \modelname is given in Algorithm~\ref{alg:training}.

\def\hitsone{\mathit{Hit}\mbox{@}1}
\def\hitsten{\mathit{Hit}\mbox{@}10}
\def\mean{\mathit{Mean}}
\def\mrr{\mathit{MRR}}

\section{Experiments}
In this section, we evaluate \modelname on two knowledge alignment tasks: cross-lingual entity alignment and zero-shot alignment.
We also conduct an experiment on cross-lingual KG completion, which aims at enhancing the traditional monolingual KG completion with cross-lingual knowledge.\par

\stitle{Dataset.} Experiments are conducted on the trilingual dataset WK3l60k, which is extracted from the subset of \mbox{DBpedia} that is highly covered by ILLs in the purpose of providing enough ground truth to evaluate the semi-supervised cross-lingual learning.
Statistics of the dataset is given in Table~\ref{tbl:statistics}.
Each language-specific version of the KG consists of 54k to 65k entities, and varies in density, which indicates the dataset to be challenging in terms of cross-lingual inconsistency and providing much larger candidate spaces than other datasets for KG embeddings that typically searches around 15k-40k entities
~\cite{yang2015embedding,sun2017cross}.
Literal descriptions covers 82\%-96\% of entities in each language.
We \mbox{extract} ILLs between English-French and English-German to train and evaluate cross-lingual entity alignment, for which we use about 20\% for training, 70\% for testing, and the rest for validation.
The proportion used for training is in accord with the estimated global completeness of ILLs in the KB~\cite{lehmann2015dbpedia}.
Meanwhile, another small set of entities with ILLs and descriptions are extracted, but are excluded from the KG structure for evaluating zero-shot alignment.

\subsection{Cross-lingual Entity Alignment}
The objective of this task is to match the same entities from different languages in KB.
The baselines we compare against include three MTransE variants that adopt different alignment techniques to model ILLs, and ITransE which employs parameter sharing for self-training.
We also adapt LM, CCA, and OT (as introduced in Section 2) to their KG equivalences.\par
\stitle{Evaluation Protocol.}
The MTransE variants, ITransE, and KGEM of \modelname are trained on the complete KG structures of two languages and the small training set of ILLs.
LM and CCA are implemented by inducing the corresponding transformations across separately trained knowledge models.
OT is implemented by enforcing MTransE-LT with an orthogonality constraint.
DEM of \modelname is trained on the entity descriptions that are covered by the current $I_{tr}$ during each iteration of co-training.
For each ILL $(e, e')$, the prediction is performed by a kNN search from the cross-lingual conversion point of $\mathbf{e}$, and record the rank of $\mathbf{e}'$ within related entities in the target language.
Following the
convention~\cite{nickel2016holographic}, we aggregate three metrics on test cases: the accuracy $\hitsone$ (\%), the proportion of ranks no larger than 10 $\hitsten$ (\%), and mean reciprocal rank $\mrr$.
All three metrics are preferred to be higher to indicate better performance.\par
Model configuration is based on the validation set.
We search the learning rate $\lambda_1$ for KGEM and other baselines among $\{0.001, 0.005, 0.01\}$, dimensionality $k_1$ in $\{50, 75, 100\}$, margin $\gamma$ in $\{0.5, 1, 2\}$, and $\alpha$ in $\{1, 2.5, 5, 7.5\}$.
For ITransE, we select the distance threshold $\delta_i$ for self-training among $\{0.5, 0.75, 1\}$.
For DEM of \modelname we select the learning rate $\lambda_2$ among $\{0.001, 0.005, 0.01\}$, dimensionality $k_2$ in $\{50, 75, 100\}$.
We fix the batch sizes $|B_t|$ for KGEM and other models, and $|B_d|$ for DEM as 1024.
The best configuration is $\lambda_1=0.005$, $k_1=50$, $\gamma=1$, $\alpha=2.5$, $\delta_i=0.75$ for all KG embedding models, and $\lambda_2=0.001$, $k_2=75$ for DEM.
For ILL proposing, we set the precision threshold $\tau$ to 0.9.
We pre-train Bilbowa based on the setting in \cite{gouws2015bilbowa} to obtain 200-dimensional word embeddings.
The multilingual entity descriptions are delimited to the first two sentences, so as to reduce some inconsistent content details.
We also remove the stop words in these descriptions, zero-pad short ones and truncate long ones to the average sequence length of 36.
Training of models is always terminated via early-stopping, and the co-training process of \modelname is terminated when either component is not able to propose ILLs for at least 1\% of the entity vocabulary.\par
{
\begin{table}[t]
\setlength\tabcolsep{2pt}
\centering
\scriptsize
\begin{tabular}{c|ccc?c|cccc}
\bhline
Data&\#En&\#Fr&\#De&ILL Lang&\#Train&\#Valid&\#Test&\#Zero-shot\\
\hline
Triples&569,393&258,337&224,647&En-Fr&13,050&2,000&39,155&5,000\\
Desc.&67,314&45,842&43,559&En-De&12,505&2,000&41,018&5,632\\
\bhline
\end{tabular}
\caption{Statistics of the Wk3l60k dataset.}\label{tbl:statistics}
\end{table}
}

{
\begin{table}[t]
\setlength\tabcolsep{2pt}
\centering
\scriptsize
\begin{tabular}{c|ccc|ccc}
\bhline
Language&\multicolumn{3}{c|}{En-Fr}&\multicolumn{3}{c}{En-De}\\
\hline
Metric&$\hitsone$&$\hitsten$&$\mrr$&$\hitsone$&$\hitsten$&$\mrr$\\
\bhline
LM&1.02&2.21&0.014&1.37&2.14&0.015\\
CCA&1.80&3.54&0.021&2.19&3.42&0.025\\
OT&20.15&25.37&0.212&11.04&19.74&0.122\\
ITransE&10.14&11.59&0.106&6.55&11.44&0.076\\
MTransE-AC&4.49&8.67&0.051&5.56&8.50&0.060\\
MTransE-TV&5.12&7.55&0.055&3.62&8.12&0.053\\
MTransE-LT&27.40&33.98&0.309&17.90&31.59&0.225\\
\hline
\modelname ($i2$)&37.70&45.01&0.405&29.80&41.66&0.322\\
\modelname ($i3$)&43.77&53.07&0.463&30.99&43.02&0.334\\
\modelname ($i4$)&46.17&54.85&0.487&32.20&44.58&0.346\\
\modelname (term)&\textbf{48.32}&\textbf{56.95}&\textbf{0.496}&\textbf{33.52}&\textbf{45.47}&\textbf{0.349}\\
\bhline
\end{tabular}
\caption{Results of cross-lingual entity alignment.}\label{tbl:entity}
\end{table}
}

{
\begin{table}[t]
\setlength\tabcolsep{2pt}
\centering
\scriptsize
\begin{tabular}{c|ccc|ccc}
\bhline
Language&\multicolumn{3}{c|}{En-Fr}&\multicolumn{3}{c}{En-De}\\
\hline
Metric&$\hitsone$&$\hitsten$&$\mrr$&$\hitsone$&$\hitsten$&$\mrr$\\
\bhline
Single-layer&0.97&1.80&0.013&0.36&2.10&0.010\\
CNN&1.19&6.91&0.036&1.28&4.63&0.019\\
GRU&18.45&27.65&0.204&11.23&24.48&0.165\\
AGRU-mono&5.08&18.27&0.096&5.03&14.90&0.085\\
AGRU-multi&26.92&44.69&0.337&19.34&45.69&0.269\\
\hline
\modelname ($i1$)&27.69&48.69&0.346&19.52&45.84&0.274\\
\modelname ($i2$)&28.82&52.58&0.350&20.37&46.35&0.279\\
\modelname ($i3$)&30.83&55.91&\textbf{0.384}&21.28&48.49&0.283\\
\modelname (term)&\textbf{30.96}&\textbf{56.93}&0.382&\textbf{21.97}&\textbf{50.02}&\textbf{0.285}\\
\bhline
\end{tabular}
\caption{Results of zero-shot alignment.}\label{tbl:zs}
\end{table}
}

{
\begin{table}[t]
\setlength\tabcolsep{1pt}
\centering
\scriptsize
\begin{tabular}{c|cc|cc|cc|cc}
\bhline
Language&\multicolumn{4}{c|}{Fr}&\multicolumn{4}{c}{De}\\
\hline
Predict&\multicolumn{2}{c|}{Tail}&\multicolumn{2}{c|}{Head}&\multicolumn{2}{c|}{Tail}&\multicolumn{2}{c}{Head}\\
\hline
Metric&$\hitsten$&$\mrr$&$\hitsten$&$\mrr$&$\hitsten$&$\mrr$&$\hitsten$&$\mrr$\\
\bhline
TransE&29.21&0.077&18.19&0.046&29.58&0.099&23.57&0.059\\
\hline
\modelnamens-mono&31.05&0.092&16.88&0.053&29.13&0.124&27.63&0.106\\
\modelnamens-cross&\textbf{37.21}&\textbf{0.139}&\textbf{22.23}&\textbf{0.093}&\textbf{34.17}&\textbf{0.134}&\textbf{31.05}&\textbf{0.143}\\
\bhline
\end{tabular}
\caption{Results of KG completion.}\label{tbl:kgc}
\end{table}
} 
\stitle{Results.} Results are reported in Table~\ref{tbl:entity}, where the results by \modelname are reported for three co-training iterations since the second iteration where KGEM is first leveraged, and for its final stage (which are respectively marked as \modelname ($i2-i4$) and \modelname (term)).
Among all baselines, MTransE-LT notably outperforms others, including other MTransE variants.
The orthogonality constraint of OT seems to be too \mbox{strict} so that it impairs the performance.
ITransE works well on aligning coherent monolingual KGs~\cite{zhu2017iterative}, but does not adapt well to the inconsistent multilingual KGs.
Without jointly adapting the monolingual vector spaces with the alignment, off-line approaches LM and CCA are left behind.
On both language settings, \modelname is able to gradually improve MTransE-LT in every iteration of co-training.
The most significant improvements happen in the first iterations, where a majority of candidate ILLs are to be proposed.
The final stages of \modelname (6$^{th}$ and 5$^{th}$ iterations of the \mbox{two} settings)
outperform the best baseline by almost doubling $\hitsone$ as well as offering significantly higher $\hitsten$ and $\mrr$.
Hence, the co-training approach of \modelname on enhancing semi-supervised entity alignment is very promising.

\subsection{Zero-shot Alignment}
This task focuses on aligning entities that do not exist in the structure of KG.
While existing KG embedding models require candidates to occur for at least once in the KG structures, \modelname is capable of dealing with zero-shot scenarios based on the representations of descriptions.
For this task, we evaluate \modelname by aligning the \emph{zero-shot set} of WK3l60k, which are excluded from the KG structures for training.
Meanwhile, we also compare the vanilla AGRU without co-training (AGRU-multi) against other encoding techniques, so as to show the effectiveness of our DEM.
These baselines include the Single-layer encoder that applies an affine layer to the averaged word embeddings of a description and the two-layer CNN with max-pooling in \cite{xie2016desc} that have been used to encode monolingual descriptions, as well as a two-layer GRU encoder without attention.
We also substitute Bilbowa with monolingual Skipgram~\cite{mikolov2013word2vec} in AGRU (AGRU-mono) so as to verify the effectiveness of incorporating multilingual word embeddings.\par
\stitle{Evaluation Protocol.} We carry forward the corresponding configurations from the last experiment to show the performance under controlled variables.
Specifically for CNN, we follow \cite{xie2016desc} to use 4-max-pooling and kernel-size of 2.
Skipgram is trained separatedly on Wikipedia dumps of two languages towards 200-dimensional word vectors for AGRU-mono.
All the baselines are trained on the ILL training set and corresponding descriptions. The results of \modelname are reported for the first three iterations and the final stage.\par
\stitle{Results.} Results in Table~\ref{tbl:zs} show that the vanilla DEM of \mbox{AGRU} outperforms the other encoders.
This also indicates that AGRU is more competent for proposing ILLs in co-training based on unseen descriptions than others.
As expected, co-training effectively leverages the zero-shot alignment with an increment of $\hitsone$ by 4.04\% and 2.63\%, as well as $\hitsten$ by 11.97\% and 4.33\% respectively on the two language settings.
The results by GRU and AGRU-mono show that self-attention and multilingual word embeddings are vital to capture the cross-lingual semantic relatedness of descriptions from the word level.
Failing to capture the sequence information, Single-layer and CNN are left behind.

\subsection{Cross-lingual KG Completion}\label{sec:completion}
Lastly, we compare the KGEM of \modelname against its monolingual counterpart TransE for KG completion, based on the sparser French and German versions of WK3l60k.
We explore with two prediction methods for \modelnamens.
\emph{Monolingual prediction} (\modelnamens-mono) aims to query the missing $h$ or $t$ of a triple $(h,r,t)$ in the same way of TransE by searching among the entities of the same language to minimize the dissimilarity function $f_r(h,t)$ (Section~\ref{sec:KGEM}).
\emph{Cross-lingual prediction} (\modelnamens-cross) provides a new method of triple completion, by converting the monolingual prediction process to the embedding space of another language, then convert the results back to the source language.
The idea of cross-lingual prediction is to leverage the traditional monolingual KG completion using a well-populated KG structure of an intermediary language given limited cross-lingual alignment.\par
\stitle{Evaluation Protocol.} We hold-out 10k French and German triples as test data. \modelname is co-trained on the rest of the training data till termination.
Cross-lingual predictions are processed in the space of English.
TransE follows the configuration of KGEM in the previous experiments, and is trained on the KG structure of each language excluding the test data.\par
\stitle{Results.} The results for $\hitsten$ and $\mrr$ are reported in Table~\ref{tbl:kgc}.
\modelnamens-mono performs at least comparably to TransE, which indicates that \modelname preserves well the characterization of monolingual KG structures.
Meanwhile, results of cross-lingual prediction prove feasibility of this new method by offering noticeably better outcomes than monolingual prediction.
Although this experiment is relatively simple, and may subject to the adequacy of knowledge in the intermediary language, this method 
opens up a new direction of future work for this task.
Moreover, suppose more languages of KGs are provided, we are interested in exploring an ensemble approach~\cite{chen2016xgboost} that interpolates multiple {\modelnamens}s on different bridges of languages to co-populate one sparse language-specific version of KG.\par

\section{Conclusion and Future Work}
In this paper, we propose 
a semi-supervised learning approach to co-train multilingual KG embeddings and the embeddings of entity descriptions for cross-lingual knowledge alignment.
Our approach \modelname effectively leverages KG embeddings for learning cross-lingual inferences on large, weakly-aligned KGs, which significantly outperforms previous models on the entity alignment task.
The zero-shot alignment task also shows the effectiveness of \modelname for improving the cross-lingual matching of entity descriptions through co-training.
Meanwhile, we 
observe that \modelname is able to enhance the traditional methods of KG completion by leveraging the information from another language.
For future work, besides the boosting approach mentioned in Section \ref{sec:completion} for cross-lingual KG completion, we seek to explore the effect of other forms of knowledge models in KGEM for encoding each language-specific KG structure.

{
\begingroup
\bibliographystyle{named}
\bibliography{ref}

\begin{thebibliography}{}

\bibitem[\protect\citeauthoryear{Bordes \bgroup \em et al.\egroup
  }{2013}]{bordes2013translating}
Antoine Bordes, Nicolas Usunier, et~al.
\newblock Translating embeddings for modeling multi-relational data.
\newblock In {\em NIPS}, 2013.

\bibitem[\protect\citeauthoryear{Bordes \bgroup \em et al.\egroup
  }{2014}]{bordes2014open}
Antoine Bordes, Jason Weston, et~al.
\newblock Open question answering with weakly supervised embedding models.
\newblock In {\em ECML-PKDD}, 2014.

\bibitem[\protect\citeauthoryear{Chen and Guestrin}{2016}]{chen2016xgboost}
Tianqi Chen and Carlos Guestrin.
\newblock Xgboost: A scalable tree boosting system.
\newblock In {\em KDD}, 2016.

\bibitem[\protect\citeauthoryear{Chen \bgroup \em et al.\egroup
  }{2017a}]{chen2017multigraph}
Muhao Chen, Yingtao Tian, et~al.
\newblock Multilingual knowledge graph embeddings for cross-lingual knowledge
  alignment.
\newblock In {\em IJCAI}, 2017.

\bibitem[\protect\citeauthoryear{Chen \bgroup \em et al.\egroup
  }{2017b}]{chen2017affine}
Muhao Chen, Tao Zhou, et~al.
\newblock Multi-graph affinity embeddings for multilingual knowledge graphs.
\newblock In {\em AKBC}, 2017.

\bibitem[\protect\citeauthoryear{Chen \bgroup \em et al.\egroup
  }{2017c}]{chen2017sampling}
Ting Chen, Yizhou Sun, et~al.
\newblock On sampling strategies for neural network-based collaborative
  filtering.
\newblock In {\em KDD}, 2017.

\bibitem[\protect\citeauthoryear{Chen \bgroup \em et al.\egroup
  }{2018}]{chen2018onto}
Muhao Chen, Yingtao Tian, et~al.
\newblock On2vec: Embedding-based relation prediction for ontology population.
\newblock In {\em SDM}, 2018.

\bibitem[\protect\citeauthoryear{Dettmers \bgroup \em et al.\egroup
  }{2018}]{dettmers2018convolutional}
Tim Dettmers, Pasquale Minervini, et~al.
\newblock Convolutional 2d knowledge graph embeddings.
\newblock In {\em AAAI}, 2018.

\bibitem[\protect\citeauthoryear{Fang \bgroup \em et al.\egroup
  }{2017}]{fangobject2017}
Yuan Fang, Kingsley Kuan, et~al.
\newblock Object detection meets knowledge graphs.
\newblock In {\em IJCAI}, 2017.

\bibitem[\protect\citeauthoryear{Faruqui and Dyer}{2014}]{faruqui2014improving}
Manaal Faruqui and Chris Dyer.
\newblock Improving vector space word representations using multilingual
  correlation.
\newblock {\em EACL}, 2014.

\bibitem[\protect\citeauthoryear{Gouws \bgroup \em et al.\egroup
  }{2015}]{gouws2015bilbowa}
Stephan Gouws, Yoshua Bengio, et~al.
\newblock Bilbowa: Fast bilingual distributed representations without word
  alignments.
\newblock In {\em ICML}, 2015.

\bibitem[\protect\citeauthoryear{He \bgroup \em et al.\egroup
  }{2017}]{he2017learning}
He~He, Anusha Balakrishnan, et~al.
\newblock Learning symmetric collaborative dialogue agents with dynamic
  knowledge graph embeddings.
\newblock In {\em ACL}, 2017.

\bibitem[\protect\citeauthoryear{Ji \bgroup \em et al.\egroup
  }{2015}]{ji2015knowledge}
Guoliang Ji, Shizhu He, et~al.
\newblock Knowledge graph embedding via dynamic matrix.
\newblock In {\em ACL}, 2015.

\bibitem[\protect\citeauthoryear{Jia \bgroup \em et al.\egroup
  }{2016}]{jia2016locally}
Yantao Jia, Yuanzhuo Wang, et~al.
\newblock Locally adaptive translation for knowledge graph embedding.
\newblock In {\em AAAI}, 2016.

\bibitem[\protect\citeauthoryear{Jozefowicz \bgroup \em et al.\egroup
  }{2015}]{jozefowicz2015empirical}
Rafal Jozefowicz, Wojciech Zaremba, et~al.
\newblock An empirical exploration of recurrent network architectures.
\newblock In {\em ICML}, 2015.

\bibitem[\protect\citeauthoryear{Kim \bgroup \em et al.\egroup
  }{2017}]{kim2017structured}
Yoon Kim, Carl Denton, et~al.
\newblock Structured attention networks.
\newblock In {\em ICLR}, 2017.

\bibitem[\protect\citeauthoryear{Koehn}{2005}]{koehn2005europarl}
Philipp Koehn.
\newblock Europarl: A parallel corpus for statistical machine translation.
\newblock In {\em MT summit}, 2005.

\bibitem[\protect\citeauthoryear{Lehmann \bgroup \em et al.\egroup
  }{2015}]{lehmann2015dbpedia}
Jens Lehmann, Robert Isele, et~al.
\newblock Dbpedia--a large-scale, multilingual knowledge base extracted from
  {Wikipedia}.
\newblock {\em Semantic Web}, 2015.

\bibitem[\protect\citeauthoryear{Lin \bgroup \em et al.\egroup
  }{2015}]{lin2015learning}
Yankai Lin, Zhiyuan Liu, et~al.
\newblock Learning entity and relation embeddings for knowledge graph
  completion.
\newblock In {\em AAAI}, 2015.

\bibitem[\protect\citeauthoryear{Mahdisoltani \bgroup \em et al.\egroup
  }{2015}]{mahdisoltani2014yago3}
Farzaneh Mahdisoltani, Joanna Biega, et~al.
\newblock Yago3: A knowledge base from multilingual {Wikipedias}.
\newblock In {\em CIDR}, 2015.

\bibitem[\protect\citeauthoryear{Mikolov \bgroup \em et al.\egroup
  }{2013a}]{mikolov2013exploiting}
Tomas Mikolov, Quoc~V Le, et~al.
\newblock Exploiting similarities among languages for machine translation.
\newblock {\em CoRR}, 2013.

\bibitem[\protect\citeauthoryear{Mikolov \bgroup \em et al.\egroup
  }{2013b}]{mikolov2013word2vec}
Tomas Mikolov, Ilya Sutskever, et~al.
\newblock Distributed representations of words and phrases and their
  compositionality.
\newblock In {\em NIPS}, 2013.

\bibitem[\protect\citeauthoryear{Nickel \bgroup \em et al.\egroup
  }{2016}]{nickel2016holographic}
Maximilian Nickel, Lorenzo Rosasco, et~al.
\newblock Holographic embeddings of knowledge graphs.
\newblock In {\em AAAI}, 2016.

\bibitem[\protect\citeauthoryear{Speer \bgroup \em et al.\egroup
  }{2017}]{speer2017conceptnet}
Robert Speer, Joshua Chin, et~al.
\newblock Conceptnet 5.5: An open multilingual graph of general knowledge.
\newblock In {\em AAAI}, 2017.

\bibitem[\protect\citeauthoryear{Sun \bgroup \em et al.\egroup
  }{2017}]{sun2017cross}
Zequn Sun, Wei Hu, et~al.
\newblock Cross-lingual entity alignment via joint attribute-preserving
  embedding.
\newblock In {\em ISWC}, 2017.

\bibitem[\protect\citeauthoryear{Thi \bgroup \em et al.\egroup
  }{2016}]{do2016facing}
Do~Thi, Ngoc Quynh, et~al.
\newblock Facing the most difficult case of semantic role labeling: A
  collaboration of word embeddings and co-training.
\newblock In {\em ACL}, 2016.

\bibitem[\protect\citeauthoryear{Wan}{2009}]{wan2009co}
Xiaojun Wan.
\newblock Co-training for cross-lingual sentiment classification.
\newblock In {\em ACL-IJCNLP}, 2009.

\bibitem[\protect\citeauthoryear{Wang \bgroup \em et al.\egroup
  }{2014}]{wang2014knowledge}
Zhen Wang, Jianwen Zhang, et~al.
\newblock Knowledge graph embedding by translating on hyperplanes.
\newblock In {\em AAAI}, 2014.

\bibitem[\protect\citeauthoryear{Xie \bgroup \em et al.\egroup
  }{2016}]{xie2016desc}
Ruobing Xie, Zhiyuan Liu, et~al.
\newblock Representation learning of knowledge graphs with entity descriptions.
\newblock In {\em AAAI}, 2016.

\bibitem[\protect\citeauthoryear{Xing \bgroup \em et al.\egroup
  }{2015}]{xing2015normalized}
Chao Xing, Dong Wang, et~al.
\newblock Normalized word embedding and orthogonal transform for bilingual word
  translation.
\newblock In {\em NAACL}, 2015.

\bibitem[\protect\citeauthoryear{Yang \bgroup \em et al.\egroup
  }{2015}]{yang2015embedding}
Bishan Yang, Wen-tau Yih, et~al.
\newblock Embedding entities and relations for learning and inference in
  knowledge bases.
\newblock In {\em ICLR}, 2015.

\bibitem[\protect\citeauthoryear{Zhang \bgroup \em et al.\egroup
  }{2014}]{zhang2014addressing}
Mi~Zhang, Jie Tang, et~al.
\newblock Addressing cold start in recommender systems: A semi-supervised
  co-training algorithm.
\newblock In {\em SIGIR}, 2014.

\bibitem[\protect\citeauthoryear{Zhu \bgroup \em et al.\egroup
  }{2017}]{zhu2017iterative}
Hao Zhu, Ruobing Xie, et~al.
\newblock Iterative entity alignment via knowledge embeddings.
\newblock In {\em IJCAI}, 2017.

\end{thebibliography}
\endgroup
}

\end{document}